\documentclass[nohyperref]{article}

\usepackage{microtype}
\usepackage{graphicx}
\usepackage{subfigure}
\usepackage{booktabs} 

\usepackage{hyperref}

\usepackage[accepted]{icml2022_new}

\usepackage{amsmath}
\usepackage{amssymb}
\usepackage{mathtools}
\usepackage{amsthm}

\usepackage[capitalize,noabbrev]{cleveref}

\theoremstyle{plain}

\theoremstyle{definition}

\theoremstyle{remark}

\usepackage[textsize=tiny]{todonotes}

\icmltitlerunning{Self-explaining Neural Network with Concept-based Explanations for ICU Mortality Prediction}

\begin{document}

\twocolumn[
\icmltitle{Self-explaining Neural Network with Concept-based Explanations for ICU Mortality Prediction}




\begin{icmlauthorlist}
\icmlauthor{Sayantan Kumar}{af1}
\icmlauthor{Sean Yu}{af3}
\icmlauthor{Thomas Kannampallil}{af3,af4}
\icmlauthor{Andrew Michelson}{af3,af5}
\icmlauthor{Philip R.O. Payne}{af3}

\end{icmlauthorlist}

\icmlaffiliation{af1}{Department of Computer Science and Engineering, Washington University in St. Louis, St. Louis, MO, USA}
\icmlaffiliation{af3}{Institute for Informatics, Washington University School of Medicine, St. Louis, MO, USA}
\icmlaffiliation{af4}{Department of Anaesthesiology, Washington University School of Medicine, St. Louis, MO, USA}
\icmlaffiliation{af5}{Department of Pulmonary Critical Care and Medicine, Washington University School of Medicine, St. Louis, MO, USA}

\icmlcorrespondingauthor{Sayantan Kumar}{sayantan.kumar@wustl.edu}

\icmlkeywords{Machine Learning, ICML}

\vskip 0.3in
]



\printAffiliationsAndNotice{}  

\begin{abstract}
Complex deep learning models show high prediction tasks in various clinical prediction tasks but their inherent complexity makes it more challenging to explain model predictions for clinicians and healthcare providers. Existing research on explainability of deep learning models in healthcare have two major limitations: using post-hoc explanations and using raw clinical variables as units of explanation, both of which are often difficult for human interpretation. In this work, we designed a self-explaining deep learning framework using the expert-knowledge driven clinical concepts or intermediate features as units of explanation. The self-explaining nature of our proposed model comes from generating both explanations and predictions within the same architectural framework via joint training. We tested our proposed approach on a publicly available Electronic Health Records (EHR) dataset for predicting patient mortality in the ICU. In order to analyze the performance-interpretability trade-off, we compared our proposed model with a baseline having the same set-up but without the explanation components. Experimental results suggest that adding explainability components to a deep learning framework does not impact prediction performance and the explanations generated by the model can provide insights to the clinicians to understand the possible reasons behind patient mortality.      
\end{abstract}

\section{Introduction}
\label{sec:intro}

Linear machine models such as logistic regression and shallow decision trees have been successfully employed in the healthcare domain due to their inherent interpretative nature and are widely used by clinicians for clinical predictive tasks such as disease diagnosis \cite{bonner2001decision, yao2005r}. However, these models can often easily perform poorly on large heterogeneous clinical datasets. On the other hand, complex deep learning models (particularly neural networks) have been shown to achieve high levels of performance in various downstream healthcare tasks because of their ability to detect complex patterns in the data \cite{lasko2013computational, kale2015causal, miotto2016deep, che2015deep}. However, the inherent complexity of black-box deep learning models makes it more challenging to explain model predictions especially for those unfamiliar, specially clinicians and healthcare providers \cite{waljee2010machine, lahav2018interpretable}. This trade-off between model performance and interpretability can lead to an important research question: how can we develop deep learning models having high predictive accuracy and at the same time can be easily interpreted and understood by health care professionals? \cite{che2016interpretable} In spite of considerable research in recent years, there exists no single widely-accepted definition of explainability or interpretability of deep learning models \cite{karim2018machine}. Model developers might be more interested in the working mechanism of the model for the purpose of debugging while clinicians might focus on understanding the rationale behind the clinical predictions obtained as model output \cite{ras2022explainable}. In order to avoid any ambiguity in the paper, we define both explainability and interpretability as the extent to which the model produces explanations about its predictions that are generally accepted as being understandable, useful, and plausible by subject matter experts (clinicians and healthcare providers) \cite{tonekaboni2019clinicians}. We will use the terms explainability and interpretability interchangeably throughout the manuscript, both referring to the same definition as above.

In the past, multiple approaches have been proposed to provide explanations for deep learning models applied on clinical data [see \cite{payrovnaziri2020explainable} for review]. However, these approaches suffer from two major drawbacks: post-hoc explanations and using raw clinical variables as units of explanation. Post-hoc explanation methods are usually motivated by the trade-off between predictive performance and interpretability. Hence, instead of modifying the existing model architecture with the risk of a lower predictive performance, post-hoc explanations accompany the original model as a separate model to provide insights about model predictions \cite{li2018deep}.  However, post-hoc explainability methods have a key issue of ownership. If an explanation for a given prediction is incorrect, it may be difficult to isolate whether the original model was responsible for this behaviour, or the explanation method generated the error. Due to these challenges, post-hoc explanations in healthcare are often not fully accepted by clinicians and hence, are not reliable enough to integrate into clinical workflow \cite{rudin2019stop}. A potential solution to this problem is developing self-explaining models which can learn self-explainable representation (i.e. no need for training separate models for explanation) and have a tight association with the prediction task via joint training (mutual benefits for both accurate prediction and accurate explanation) \cite{alvarez2018towards}.


\begin{figure*}[ht]
\vskip 0.2in
\centerline{\includegraphics[width=12cm]{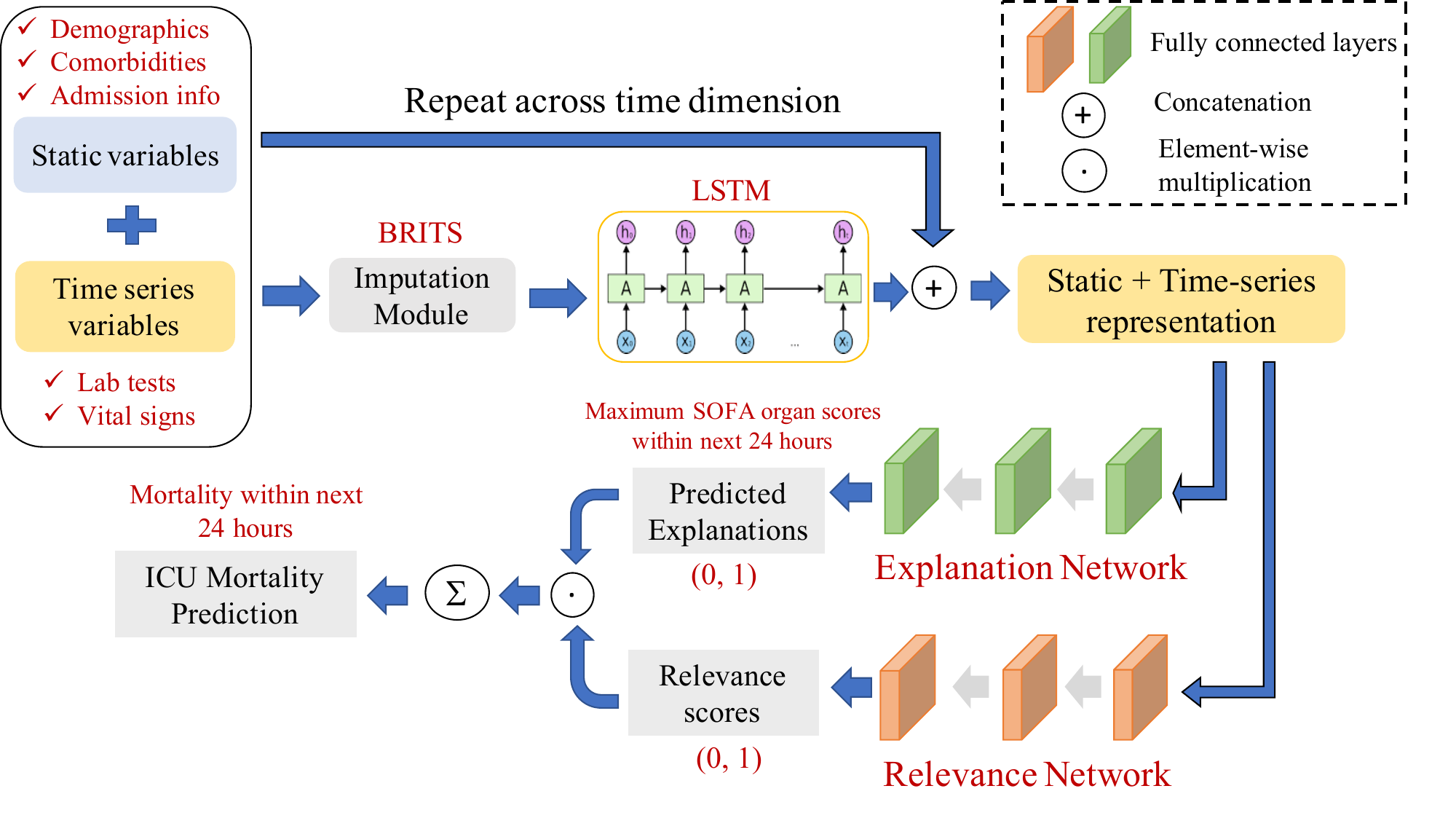}}
\caption{The time-series variables (lab tests and vital signs) are passed through an imputation module BRITS and the imputed variables are propagated through recurrent (LSTM) layers to yield a latent (hidden) representation. The final latent representation, formed by concatenating the static variables (repeated across time dimension) is passed through a series of fully connected neural network layers to generate predicted explanations or auxiliary scores (predicted maximum SOFA organ scores within next 24 hours). The relevance scores (weights/significance) of each concept are generated through a separate set of fully connected layers. The weighted sum of the explanations and relevance scores can be used to yield the predicted probabilities of mortality within next 24 hours.}
\label{fig:pipeline}
\vskip -0.2in
\end{figure*}


Another challenge of existing explainability approaches in healthcare is to explain the model predictions in terms of the raw clinical features used as input to deep learning models. Weights are assigned to individual features highlighting their contribution (importance) towards model prediction, e.g. logistic regression \cite{thomas2008measuring}, saliency maps \cite{kindermans2019reliability} and Shapley explanations \cite{kumar2020problems}. However, using raw features (e.g. pixels in medical images) as units of explanation is often difficult for human interpretation and can lead to unstable explanations that are sensitive to noise or perturbations in the input. To mitigate some of these challenges, we can operate on higher-level features or feature intermediates, which can be referred to as high-level concepts derived from a combination of raw features. These high-level concepts can be understood as aggregated knowledge which clinical experts often rely on to make decisions. For example, in medical imaging, high-level concepts driven by expert knowledge such as tissue ruggedness or elongation are strong predictors of cancerous tumours and can be the natural "units" of explanation for doctors to make their diagnosis \cite{koh2020concept}. 

In this work, we aim to address both the above-mentioned challenges of generating deep learning explanations when applied in the healthcare setting. We designed a self-explaining deep learning framework using the expert-knowledge driven clinical concepts or intermediate features derived from raw clinical variables as units of explanation. Our proposed framework uses Sequential Organ Failure Assessment (SOFA) scores as high-level concepts to explain patient mortality in the ICU. SOFA score is a composite score derived from the organ-based sub scores that measure an individual patient's degree of organ dysfunction using raw clinical data and can be tracked over time. The SOFA score is a well-validated metric associated with ICU  mortality outcome and functions as an aggregated assessment of clinical status thus, these scores can become ideal candidates to be units of explanation and provide insights into the reasoning behind mortality prediction.

In order to address the performance-interpretability trade-off, we designed our framework to be intrinsically interpretable by jointly predicting and explaining the predictions. We leverage the benefits of multi-task learning to predict the explanations or concepts as supervised auxiliary tasks and utilize the predicted explanations to predict the final outcome, both in a joint end-to-end training setting. The self-explaining nature of our proposed model comes from the relevance scores we generate for each concept within the framework itself without any additional training effort. The relevance scores can provide insights into the question of "why did the patient die?" and quantifies the importance of each concept (SOFA organ system score) in deciding the final mortality outcome. Our contributions can be summarized as follows:
\begin{itemize}
    \item To the best of our knowledge, our proposed approach is the first study which uses both supervised high-level clinical concepts and intrinsically developed model explanations in the context of predicting a clinical outcome. 
    \item We test our proposed approach on a publicly available longitudinal Electronic Health Records (EHR) dataset for predicting anticipated mortality within the next 24 hours at each point in the ICU trajectory.
    \item We answer the following research questions through our experiments: (a) Does adding explainability components to a deep learning framework affect its prediction performance (interpretability-performance trade-off), (b) Are the predicted explanations grounded in terms of expert domain knowledge? and (c) Do the explanations generated by our method provide insights into patient mortality? 
\end{itemize}

\section{Related Work}

In the following section, we briefly describe the studies related to our work. We divide the related research works into 2 broad categories: (i) interpretability methods for neural networks, describing both existing post-hoc and intrinsically explainable neural network based frameworks and (ii) methods which provide explanations through high-level concepts or prototypes.

\subsection{\textbf{Interpretability methods for neural networks}}

There exist several interpretability methods for neural networks which mostly focus on gradient and perturbation-based methods mentioned here \cite{bach2015pixel, shrikumar2017learning, simonyan2014deep, sundararajan2017axiomatic}. All these methods do not modify the existing architecture, but explain model predictions in terms of the importance values (weights) of the input features or sensitivities of the inputs with change in the outcome. Our proposed approach differs from the above studies in both the units of explanation - high level concepts instead of raw features and how they are used, relying on the relevance scores produced intrinsically by the model, eliminating the need for additional computation. Studies focusing on intrinsic explanations associated with medical prediction tasks have mostly focused on attention mechanism to find a set of input variables with the most relevant information to the prediction task. Attention has been used to (i) highlight when input features have influenced predictions of clinical events of ICU patients \cite{kaji2019attention}, (ii) design an interpretable acuity score based on DeepSOFA that can evaluate a ICU patient's severity of illness \cite{shickel2019deepsofa} and (iii) learn a representation of EHR data that captures the relationships between clinical events for each patient (Patient2Vec) \cite{zhang2018patient2vec}. Choi et al. \cite{choi2016retain} designed a reverse time attention model (RETAIN) which mimic the behaviour of a medical professional and incorporate sequential information. Kwon et al. \cite{kwon2018retainvis} developed a visually interpretable deep learning framework for heart failure based on RETAIN. Unlike our proposed approach, all the above attention based methods use individual clinical features as units of explanation and do not use high level feature concepts or any kind of aggregated knowledge.  

\subsection{\textbf{Explanations through concepts and prototypes}} A recent study \cite{alvarez2018towards} used a self- explainable neural network, learning concepts in an unsupervised way to explain model predictions. We adopt a similar approach in the context of clinical predictive modelling and design high-level clinical concepts which can be learned in a supervised way driven by medical expert-knowledge. Li et al. \cite{li2018deep} proposes an interpretable deep learning framework whose predictions are based on the similarity of the input to a small set of prototypes which are learned during training. Kim et al. \cite{kim2018interpretability} follows a post-hoc approach of learning interpretable concepts to explain model predictions. Their proposed model learns concept activation vectors representing human friendly concepts of interest and the directional derivatives along these vectors can be used to estimate the sensitivity of the predictors with respect to semantic changes in the direction of the concept. Mincu et al. \cite{mincu2021concept} extended the TCAV concept to longitudinal EHR data. Their approach differs from ours is that the contribution of the concept to the primary task can only be determined globally (through aggregation over multiple samples) and not locally (for each patient). Our proposed approach allows local explanations through the relevance scores that can explain the mortality of each patient.

\section{Proposed Framework}

In this section, we describe our proposed explainability framework (Figure \ref{fig:pipeline}). Our method has a multi-task setting where the explanations or high-level concepts are predicted as auxiliary tasks and the predicted auxiliary scores are used to generate the final prediction output. Figure \ref{fig:pipeline} shows all the components of our proposed framework in detail.

Let $X$ be a multivariate time series of a particular patient which can be represented by a sequence of T observations $\{x_t\}$, where $t = 1, 2,...T$. We denote the set of time invariant or static features of that patient as $\{x_s\}$. The time-series features are passed through an imputation algorithm named BRITS which imputes the missing values according to recurrent dynamics \cite{cao2018brits}. In BRITS, the imputed variables are regarded as variables in a bidirectional RNN graph. Hence the missing values get delayed gradients in both forward and backward direction, which makes the estimation of missing values more accurate. The imputation is supervised by the imputation loss function $L_{impute}$ which can be calculated by the Mean Absolute Error (MAE) between the actual and imputed variables, as shown in Equation \ref{eq:impute_loss}. The imputed variables $\{\Tilde{x_t}\}$ (Equation \ref{eq: RITS}) are then passed through a series of recurrent Long Short Term Memory (LSTM) layers to yield a latent representation $\{h_t\}$. We define a mask $m_t$ (Equation \ref{eq:mask_define}) which keeps track of the time points where a particular feature variable is missing. In case of missing time points, we replace the missing values in $x_t$ with the corresponding values in $x_{impute}$ (Equation \ref{eq:mask}). The latent representation obtained as LSTM output is then concatenated with the time-invariant variables $\{x_s\}$, repeated across time dimension to form the final latent representation f (Equation \ref{eq:concat})

\begin{gather}
m_{t} =
    \begin{cases}
      0 & \text{$x_{t}$ is not observed at timestep t} \\ \label{eq:mask_define}
      1 & \text{otherwise} 
    \end{cases} \\
    x_{impute}^t= BRITS(\{x_{t}\} \label{eq: RITS}\\
    {\Tilde{x_t}} = m_t * x_t + (1-m_t)*x_{impute}^t \label{eq:mask} \\
    L_{impute} = \sum_{i=1}^{T}|x_t-x_{impute}^t| \label{eq:impute_loss} \\
    f = \Tilde{x_t} \oplus x_s \label{eq:concat}
\end{gather}

The concatenated representation f is then passed through a series of fully-connected layers, followed by a sigmoid activation function to generate the auxiliary layer output or predicted explanations. The auxiliary task performance is supervised by the auxiliary loss $L_{aux}$, which is calculated by the Mean Squared Error (MSE) between the predicted ($y_{aux}^{pred}$) and ground truth ($y_{aux}$) labels (Equation \ref{eq:L_aux}). The number of auxiliary tasks or explanation units is a pre-defined value based on the number of expert-knowledge driven medical concepts. We also generate relevance scores for each concept, which is estimated by passing the concatenated representation through a separate set of fully connected layers, followed by the sigmoid activation function to make the relevance scores range between 0 and 1. The relevance score of a concept at each time point signify weight or contribution of that concept in deciding the final predicted probability. Higher weights (close to 1) indicate greater contribution.  

The predicted auxiliary values $exp_j$ can be combined with the corresponding relevance scores $\alpha_j$ to generate the final probability $y_{pred}$, where j denotes the $j$-th explanation and N denotes the number of explanations (Equation \ref{eq:y_mort}). The final loss function is a weighted sum of 3 supervised losses: $L_{mort}$ (Equation \ref{eq:L_mort}), binary cross-entropy (CE) loss between ground truth $y$ mortality label $y_{pred}$ and predicted mortality $\hat{y_t}$, the auxiliary loss $L_{aux}$ and imputation loss $L_{impute}$ (Equation \ref{eq:L_total}). Since the imputation loss is dominated by more frequent variables, there exists a high degree of loss imbalance between the imputation loss and the primary and auxiliary losses, which can lead to negative loss transfer and low prediction performance \cite{li2020deepalerts}. To address this challenge, we selected higher weights for the primary and auxiliary loss ($\lambda_1 = 1, \lambda_2 = 10$) compared to the imputation loss ($\lambda_3 = 0.001$). 

\begin{gather}
L_{aux} = \sum_{i=1}^{T}(y_{aux}-y_{aux}^{pred})^2  \label{eq:L_aux} \\
y_{pred} = \sigma(\sum_{j=1}^{N} (\alpha_j * exp_j)) \label{eq:y_mort} \\
L_{mort} = -{(y\log(y_{pred}) + (1 - y)\log(1 - y_{pred}))} \label{eq:L_mort} \\
L = \lambda_{1} L_{mort} + \lambda_{2} L_{aux} + \lambda_{3} L_{impute} \label{eq:L_total}
\end{gather}


\section{Experiments}

We aim to answer the following research questions through our experiments: Does adding explainability components to a deep learning framework affect its prediction performance (interpretability-performance trade-off), (b) Are the predicted explanations grounded in terms of expert domain knowledge? and (c) Do the explanations generated by our method provide insights into patient mortality? In the remainder of this section, we will describe the dataset, details about feature engineering and pre-processing, the baseline methods for comparison and the implementation details. 

\subsection{Dataset and Experimental design}

\subsubsection*{Study participants}

We conduct our experiments on the publicly available Medical Information Mart for Intensive Care IV (MIMIC-IV v0.4) database. MIMIC-IV comprises of more than a decade worth of de-identified ICU patient records, including vital signs, laboratory and radiology reports, and therapeutic data, from patients admitted to the Intensive Care Units of the Beth Israel Deaconess Medical Centre in Boston, Massachusetts, and is freely available for research \cite{johnson2016mimic}. We established our cohort based on the following eligibility criteria: (i) ICU stays longer than 48 hours and (ii) patients older than 18 patients at the time of admission. Only the first admission was considered in case of multiple admissions to the ICU. Our cohort comprises of clinical data of 22,944 ICU admissions between 2008-2019, of which 2043 (8.9\%) experienced in-hospital mortality. The median age of adult patients (age >= 18 years old) is 67 years (IQR: 56-78 years) and the median length of stay (LOS) in the ICU is 84.3 hours (IQR: 62.3 - 132.7 hours).

\subsubsection*{Feature variables and preprocessing}

For each patient, we extracted 24 static or time-invariant features such as demographics (age, gender, race, ethnicity), admission diagnoses and comorbidity information and 87 time-series features which includes laboratory test results and vital signs. Feature pre-processing of time-series variables steps include clipping the outlier values to the 1st and 99th percentile values and standardization using the RobustScalar package from sklearn \cite{bisong2019introduction}. Time-varying variables were aggregated into hourly time buckets using the median for repeated values. All categorical features were one-hot encoded. Missing values in time-series variables were imputed using the BRITS algorithm, a state-of-the-art imputation based on recurrent dynamics \cite{cao2018brits}.

\subsubsection*{High-level concepts: Units of Explanation}

In our work, we used the different components of the Sequential Organ Failure Assessment (SOFA) scores as our units of explanation. SOFA is a widely-used score validated by clinicians and used for assessing severity of illness measuring the extent of a the person's organ function or failure. The score is based on six subscores, one each for the six organ systems: respiratory, cardiovascular, hepatic, coagulation, renal and neurological systems,  allocating a score of 0-4 with high scores indicating severe organ conditions \cite{ferreira2001serial}. Most in-hospital ICU deaths are preceded by signs of organ failure; thus, these scores can provide insights into the reasoning behind mortality prediction \cite{yu2021comparison}. At each timepoint within a patient's ICU trajectory, the maximum SOFA sub-score (for each of the six organ systems) within the next 24 hours were used as ground truth labels to supervise the predicted auxiliary score or explanations. In our model, the predicted SOFA organ scores form the six units of explanation, which when combined with the relevance scores can be used to predict the final mortality. The model predicts mortality at each timepoint as the mortality risk or the probability of patient mortality within next 24 hours. The SOFA organ scores for the six organ systems were calculated based on a set of standard rules proposed in \cite{ferreira2001serial,  lambden2019sofa}. All the SOFA organ scores were scaled between 0 and 1 by the MinMax scaling package from sklearn \cite{bisong2019introduction}.

\begin{figure}[ht]
\vskip 0.2in
\begin{center}
\centerline{\includegraphics[width=\columnwidth]{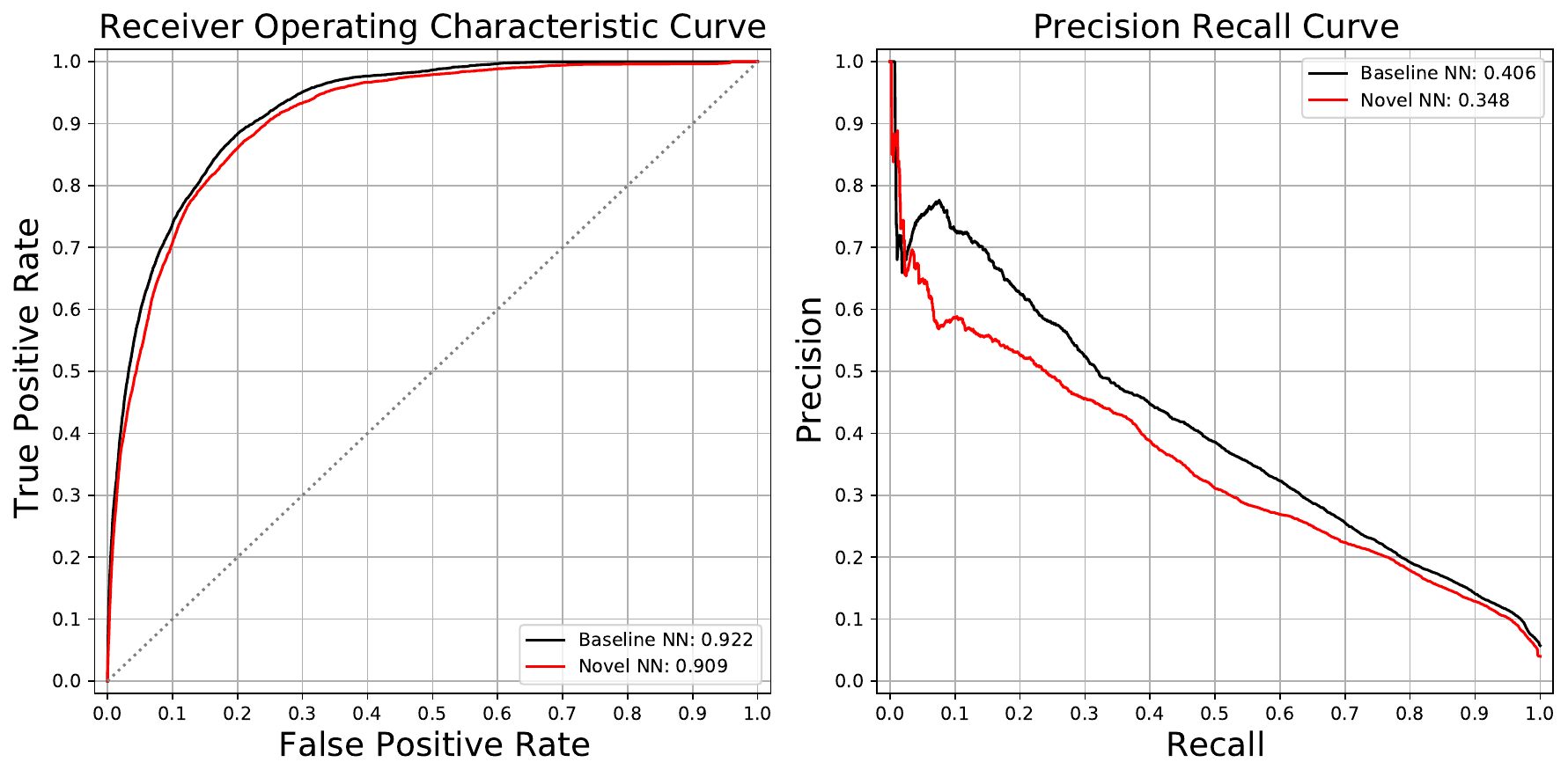}}
\caption{AUROC (AUROC (Area Under the Receiving Operating Characteristics Curve) and AUPRC (Area Under the Precision Recall curve for both our proposed (Novel NN) and baseline method (baseline NN). Both the proposed and baseline model have similar AUROC and AUPRC values indicating that adding explainability components does not impact prediction performance.}
\label{fig:ROC}
\end{center}
\vskip -0.2in
\end{figure}

\subsection{Baseline method and Implementation details}

In order to validate our hypothesis that adding explainability components to the network does not affect prediction performance, we compared our proposed approach with a baseline model to analyze the performance-interpretability trade-off. The baseline model has the same set up as our proposed model but without the components generating the auxiliary and relevance scores. In the baseline model framework, the time-series variables are passed through the imputation module BRITS and the imputed variables are propagated through recurrent (LSTM) layers to yield a latent (hidden) representation. The final latent representation, formed by concatenating the static variables (repeated across the time dimension) is passed through a series of fully connected neural network layers, followed by a sigmoid activation function to generate the final mortality probability (mortality within next 24 hours). 

Both the proposed and baseline models are trained with the same set of static and time-series input variables. The dataset was split into training, validation and test set (70:15:15), with the validation set used for early stopping. For a fair comparison, both the proposed and baseline model were trained using the same set of parameter configurations as follows: Adam optimizer with learning rate $= 0.001, \beta_1 = 0.9, \beta_2 = 0.999$ and L2 regularization factor = $0.0001$ and the number of LSTM layers was set to 3, each of dimension 128. For the imputation module BRITS, we used the same parameter configurations used in the original paper (2 recurrent layers with size 256 each). In the proposed model, the concatenated representation of static and time series variables are fed through 2 separate sets of 3 fully connected layers of size 256, 128, 64 for generating the auxiliary and relevance scores. Both the models were trained for 500 epochs with batch size 128 and dropout rate = 0.5.

\begin{figure}[ht]
\vskip 0.2in
\begin{center}
\centerline{\includegraphics[width=\columnwidth]{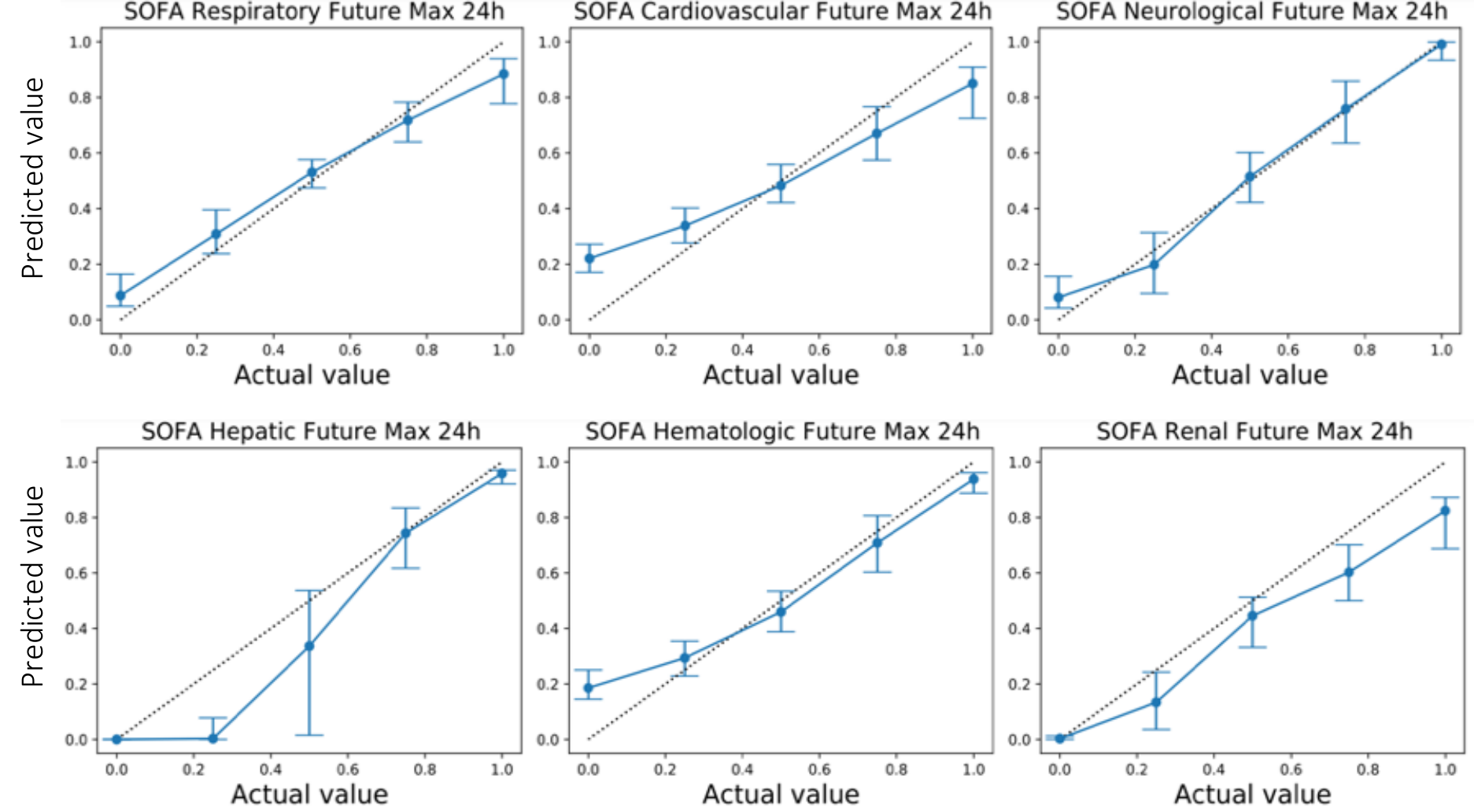}}
\caption{Performance of our proposed model on the auxiliary tasks, showing actual and predicted values of the auxiliary tasks corresponding to the maximum SOFA organ system scores within 24 hours. The dotted line in each figure represents the ideal scenario where the predicted and actual values are same.}
\label{fig:aux_MSE}
\end{center}
\vskip -0.2in
\end{figure}

\begin{figure*}[ht]
\vskip 0.2in
\begin{center}
\centerline{\includegraphics[width=10cm]{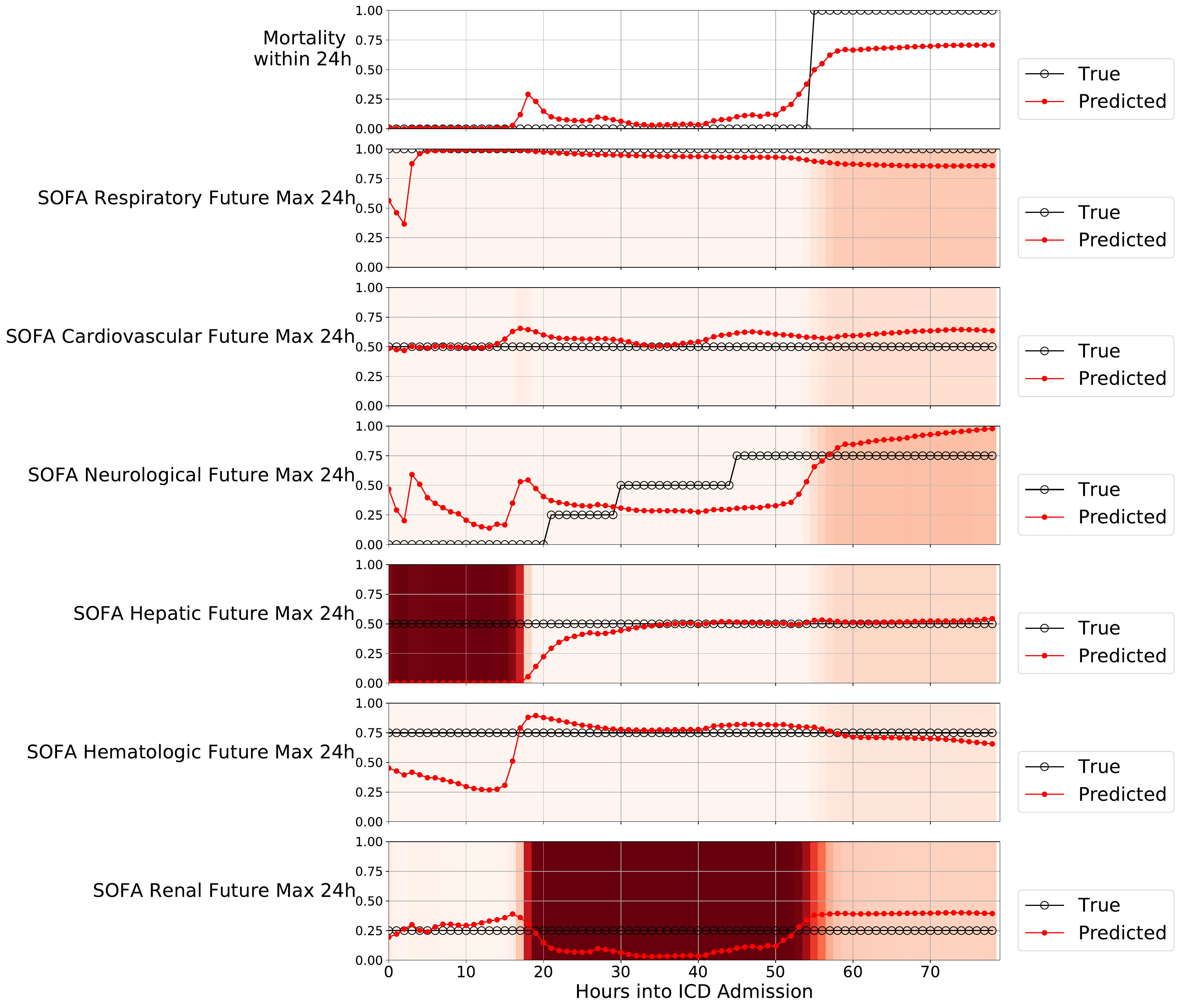}}
\caption{Figure showing how the predicted explanations (maximum SOFA organ scores within next 24 hours) and their corresponding relevance scores at each timepoint throughout the longitudinal trajectory of a particular patient in the ICU. Both the explanations and relevance scores range between 0 and 1. The topmost plot represents mortality while the following points represent the explanations. The x-axis represents time (hours into ICU admission) and the plotted values represent either the ground truth label (black) or the predicted value (red).The highlighting within each row indicates the weight/importance (relevance score) given to the explanations where dark hues corresponds to higher relevance given to a particular organ system at a specific time point.}
\label{fig:attention}
\end{center}
\vskip -0.2in
\end{figure*}


\section{Results}

Figure \ref{fig:ROC} shows the AUROC (Area Under the Receiving Operating Characteristics Curve) and AUPRC (Area Under the Precision Recall Curve) for both our proposed approach and baseline method without the explainability components (auxiliary scores and relevance scores). Both models have low AUPRC, caused by the low prevalence of mortality, or class imbalance; in our cohort, only 2043 (8.9\%) out of 22,944 ICU admissions experienced in-hospital mortality.  Both the proposed and baseline model have similar AUROC ($0.923[0.915-0.947]$ vs $0.909[0.895-0.928], p = 0.676$) and AUPRC ($0.224[0.205-0.247]$ vs $0.227[0.198-0.237], p = 0.823$) values without any statistically significant differences at 95\% confidence interval. This observation addresses our 1st research question regarding performance-interpretability trade-off and indicates that adding intrinsic explainability components does not impact prediction performance of our proposed method. 

Figure \ref{fig:aux_MSE} shows the performance of our proposed model on the auxiliary tasks corresponding to the six SOFA organ systems. For each of the organ systems, the SOFA scores ranging from 0-4 in increasing order of severity are scaled to (0,1). The results suggest that the model performs relatively well in predicting scores for all the organ systems. The decent performance metrics on the auxiliary tasks demonstrate that the predicted auxiliary scores are similar to the actual SOFA organ system scores. In other words, the predicted explanations are grounded in terms of expert knowledge and can act as reliable indicators to provide insights into potential reasons behind the final mortality outcome. 

Figure \ref{fig:attention} shows the explanation relevance visualization of the longitudinal trajectory of a single patient who died 80 hours after ICU admission. Here we can observe how the predicted explanations (maximum SOFA organ scores within next 24 hours) and their corresponding relevance scores vary at each timepoint throughout the longitudinal trajectory of the patient in the ICU. We can observe that the model initially pays the maximum importance (weight) to the anticipated SOFA hepatic dysfunction till t = 20 hours, followed by anticipated SOFA kidney (renal) dysfunction. As the predicted probability of mortality rises, the model is shown to pay more importance to anticipated respiratory, neurological, hepatic and renal organ failure, highlighting their contribution towards mortality. The explanations along with the relevance scores in Figure 4 can help clinicians understand the health status of a patient throughout the duration of the ICU stay and analyze possible reasons for mortality (if applicable).

\section{Discussion}

Our aim was to design a self-explaining deep learning framework using the expert-knowledge derived clinical concepts such as SOFA organ system scores as units of explanation for predicting longitudinal mortality in the Intensive Care Unit (ICU) setting. The self-explaining nature of our proposed model comes from the relevance scores we generate for each concept which can provide insights into the reasoning behind patient mortality. We leverage the benefits of multi-task learning to predict the explanations or concepts as supervised auxiliary tasks and utilize the same explanations to predict the final outcome, both in a joint end-to-end training setting, with the aim of producing both accurate predictions and accurate explanations.

\subsection{Interpretability - performance trade-off}

Interpretability and performance currently stand in apparent conflict in deep learning. Most relevant literature have focused on using a separate post-hoc model for explaining the predictions of a deep learning framework. These methods focus on improving the prediction performance and hence do not modify the existing architecture. In our work, we analyze the interpretability-performance trade-off by designing a framework that is jointly trained to both predict and explain in an end-to-end manner, without any additional training effort. We leveraged the benefits of a multi-task learning by predicting the explanations as auxiliary scores and using the predicted auxiliary scores to generate the final outcome. Both the auxiliary tasks and the final task are learned via joint training in a supervised manner, thus improving the performance on both the final and auxiliary tasks. Our AUROC and AUPRC results demonstrate that model performance does not change significantly even if we add an intrinsic explainability component within the main architecture. We believe that this can motivate future research on designing deep learning frameworks which can predict and explain at the same time without any cost in prediction performance. 

\subsection{Pre-defining SOFA organ scores as units of explanation}

A common approach for generating explanations for a deep learning model when used in the clinical context is to use the raw clinical variables, understanding which of them are correlated with the final prediction and in what capacity. However, for a large number of input clinical features, it often becomes difficult for the clinicians to comprehend the exact reasoning behind the final clinical outcome if we just provide them with a list of features with high weights or importance towards the final prediction. Subject matter experts tend to rely on some intermediate knowledge schemas which make them help decisions. These can be understood as some kind of aggregated knowledge or high-level interpretable concepts, which are derived from the raw clinical variables and more stable with respect to noise and input perturbations. SOFA organ system scores can be a suitable candidate for high-level interpretable concepts that are not only derived from a combination of feature variables but also widely used by clinicians to analyze the risk of patient mortality in the ICU. In our work, we train our model to learn these interpretable concepts from raw clinical variables in a supervised manner through a framework of auxiliary tasks. Our results demonstrate that the predicted explanations are grounded in terms of expert knowledge and can act as reliable indicators to explain the reasoning behind the final mortality outcome. Hence a more natural unit of explanation would be "the patient died due to organ failures in the cardiovascular system".

\subsection{Generalizability: Application of proposed framework to other clinical problem set-up}

Our goal was to design a generalizable interpretable network that can generate both predictions and explanations in any kind of clinical prediction problem. As a starting point, we used ICU mortality prediction as a single use case in our experiments, where the SOFA organ system scores were representatives of the high-level interpretable concepts used as units of explanation. However, our framework can be applied in other clinical domains too. Since we use expert-knowledge driven aggregated knowledge or high-level intermediate concepts in a supervised setting, our only assumption is that the intermediate concepts need to be pre-defined. Clinical experts have designed intermediate constructs or aggregated knowledge for different clinical problems to make decisions. An example of intermediate construct is Clinical Dementia Rating which is derived from individual cognitive performance tasks for predicting Alzheimer's Disease progression. Due to the ever-increasing volume of EHR data, clinicians often rely on similar existing intermediate knowledge derived from clinical variables for their diagnosis. Hence it is possible to obtain the intermediate knowledge as concepts specific to each clinical prediction problem. Thus our proposed framework is generalizable to work on other informatics problems too.

\subsection{Limitations and scope for future work}

In this work, we applied our proposed explainability framework on a single dataset and a specific clinical prediction problem. Our immediate next step is to test the generalizability of our proposed interpretability framework on multiple EHR datasets. One of the fundamental assumptions of our approach is that the intermediate concepts need to be pre-defined in the context of the clinical problem. Hence, our explainability framework is not applicable in cases where there exists no available expert knowledge for that particular problem. As a next step, we plan to have our model learn the interpretable concepts in an unsupervised manner without relying on expert knowledge and apply them to more complex domains such as medical imaging, speech recognition and clinical natural language processing. 

A potential limitation of SOFA organ system scores is that, it only evaluates a few organs and sometimes markers of damage to one organ actually represent markers of damage to another. For example, bilirubin can be a sign of haemolysis or gallbladder issues and also of liver failure \cite{yu2021comparison}. Also, chronic/underlying causes that may impact SOFA scores are not  accounted for in many of the studies used for clinical validation. However, we re-emphasize that our goal is to analyze if it's possible to generate plausible explanations using an example of expert-knowledge driven intermediate concepts. In our work, SOFA is used as an example test case only which satisfied the characteristics of intermediate aggregated knowledge that can be used by clinicians to make decisions. 
\section{Conclusion}

In this work, we designed a self-explaining deep learning framework with expert-knowledge driven clinical concepts as the units of explanation.  The self-explaining nature of our proposed model comes from generating both explanations and predictions via joint training. We tested our proposed approach on the MIMIC IV EHR dataset for predicting patient mortality in the ICU using SOFA organ system scores as the intermediate high-level concepts. The explanations generated by our model along with the corresponding relevance scores help clinicians monitor the health status of a patient and provide insights into possible reasons behind patient mortality. Our experiments  analyzed the interpretability-performance trade-off and demonstrate that model performance does not change significantly even if we add an intrinsic explainability component within the main architecture itself. Future work include testing the generalizability of our proposed interpretability framework on multiple EHR datasets and learn the interpretable concepts in an unsupervised manner without relying on expert knowledge.

\nocite{langley00}

\bibliography{references}
\bibliographystyle{icml2022}


You can have as much text here as you want. The main body must be at most $8$ pages long.
For the final version, one more page can be added.
If you want, you can use an appendix like this one, even using the one-column format.

\end{document}